\documentclass{article}

\PassOptionsToPackage{numbers, compress}{natbib}
\usepackage[preprint]{cml4impact_2022}

\usepackage[utf8]{inputenc} % allow utf-8 input
\usepackage[T1]{fontenc}    % use 8-bit T1 fonts
\usepackage{hyperref}       % hyperlinks
\usepackage{url}            % simple URL typesetting
\usepackage{booktabs}       % professional-quality tables
\usepackage{amsfonts}       % blackboard math symbols
\usepackage{nicefrac}       % compact symbols for 1/2, etc.
\usepackage{microtype}      % microtypography
\usepackage{xcolor}         % colors

\usepackage{algorithm}
\usepackage{algorithmic}

\usepackage{nicefrac}
\usepackage{microtype}
\usepackage{graphicx}
\usepackage{caption}
\usepackage{subcaption}
\usepackage{tikz-cd}
\usetikzlibrary{arrows}
\usepackage{comment}
\usepackage{amssymb}
\usepackage{tikz}
\usepackage{tikz-qtree,tikz-qtree-compat}
\usetikzlibrary{positioning}

\usepackage{amsmath}
\usepackage{amsthm}
\usepackage{tikz}
\usepackage{tikz-qtree,tikz-qtree-compat}

\makeatletter
\newtheorem*{rep@theorem}{\rep@title}
\newcommand{\newreptheorem}[2]{%
\newenvironment{rep#1}[1]{%
 \def\rep@title{#2 \ref{##1}}%
 \begin{rep@theorem}}%
 {\end{rep@theorem}}}
\makeatother

\newtheorem{theorem}{Theorem}
\newreptheorem{theorem}{Theorem}

\usetikzlibrary{positioning}
\newtheorem{definition}[theorem]{Definition}

\usepackage{stackengine}
\def\delequal{\mathrel{\ensurestackMath{\stackon[1pt]{=}{\scriptstyle\Delta}}}}

\title{Learning Probabilities of Causation from Finite Population Data}

\author{%
  Ang Li, Song Jiang, Yizhou Sun, Judea Pearl\\
  Department of Computer Science\\
  University of California Los Angeles\\
  Los Angeles, CA 90095 \\
  \texttt{\{angli,songjiang,yzsun,judea\}@cs.ucla.edu} \\
}

\begin{document}

\maketitle

\begin{abstract}
This paper deals with the problem of learning the probabilities of causation of subpopulations given finite population data. The tight bounds of three basic probabilities of causation, the probability of necessity and sufficiency (PNS), the probability of sufficiency (PS), and the probability of necessity (PN), were derived by Tian and Pearl. However, obtaining the bounds for each subpopulation requires experimental and observational distributions of each subpopulation, which is usually impractical to estimate given finite population data. We propose a machine learning model that helps to learn the bounds of the probabilities of causation for subpopulations given finite population data. We further show by a simulated study that the machine learning model is able to learn the bounds of PNS for $32768$ subpopulations with only knowing roughly $500$ of them from the finite population data.
\end{abstract}

\section{Introduction}
The probability of causation is a crucial concept that belongs to the third ladder of causality defined by Pearl \cite{pearl2018book} and plays a significant role in modern decision-making. The applications include the areas of marketing, political science, and health science. For example, using a linear combination of the probabilities of causation, Li and Pearl defined the benefit function of the unit selection problem, which is considered to be a revolution of the traditional A/B test heuristics \cite{li:pea22-r517, li:pea19-r488, li2022unit}. Personalized decision-making, for another example, has been demonstrated that it should consider the probabilities of causation \cite{mueller:pea-r513}. The label of a machine learning algorithm should also combine the probabilities of causation terms to capture the counterfactual behavior of the desired task \cite{li2020training}.

Using structural causal model (SCM) \cite{galles1998axiomatic,halpern2000axiomatizing,pearl2009causality}, Pearl \cite{pearl1999probabilities} first defined three basic probabilities of causation (i.e., PNS, PN, and PS). These probabilities of causation were then bounded tightly by Tian and Pearl \cite{tian2000probabilities} using Balke's programming \cite{balke1995probabilistic}. The theoretical proof of those bounds was provided at \cite{li:pea19-r488,li2022unit}. After that, researchers started to use covariate information and the causal structure to narrow the bounds of the above probabilities of causation \cite{dawid2017, pearl:etal21-r505}. Li and Pearl \cite{li:pea-r516} also presented extended studies of the probabilities of causation with nonbinary treatment and effect.

All listed works above are focused on providing bounds of the probabilities of causation on a specific population and require the experimental and observational distributions of the population. If the population can be divided into subpopulations by some observed characteristics, then the estimations of the probabilities of causation for each subpopulation require the experimental and observational distributions of each subpopulation. Consider the following example: an online music provider wants to increase its user subscription rate by sending gifts to new subscribers. Therefore, based on customer characteristics such as income, age, and usage, the company wants to identify customers who are likely to subscribe if and only if they received the gift. The gifts are value earphones; thus, the music provider prefers that these gifts be made only to those identified customers. Therefore, the music provider wants to know which kind of characteristics contains more desired customers. The quantities that the music provider wanted are then the PNS of subpopulations for each set of customer characteristics. If we want to apply Tian-Pearl's PNS bounds \cite{tian2000probabilities}, we need the experimental and observational distributions of each set of customer characteristics. However, the data available are usually finite many for the whole population; therefore, 1) it is impractical to estimate each distribution because the number of subpopulations is large; 2) some of the subpopulations are very rare (i.e., they appeared with a low probability) or even with no data associated; 3) we are not able to forecast the new coming subpopulations.

In this work, we are considering binary treatment and effect and will focus on PNS. It is not hard to extend to the other probabilities of caution. We also classify individual behavior into four response types, labeled complier, always-taker, never-taker, and defier \cite{angrist1996identification,balke1997bounds,li:pea19-r488}. Compliers are individuals who would respond positively if encouraged and negatively if not encouraged (i.e., PNS is the fraction of compliers). Always-takers are individuals who always respond positively whether or not they are encouraged. Never-takers are individuals who always respond negatively whether or not they are encouraged. Defiers are individuals who would respond negatively if encouraged and positively if not encouraged. We assume that the response type of an individual is determined by his characteristics (observed or unobserved characteristics); therefore, we propose a machine learning framework that is able to output the lower and upper bounds of PNS for each subpopulation given finite population data (i.e., learn the relations between the characteristics and the response types).

\section{Preliminaries}
\label{related work}
Here, the definitions of three basic probabilities of causation are reviewed \cite{pearl1999probabilities}. We follow the language of counterfactuals in structural causal model in \cite{galles1998axiomatic,halpern2000axiomatizing}. 

We use $Y_x=y$ to denote the basic counterfactual sentence ``Variable $Y$ would have the value $y$, had $X$ been $x$''. If not specified, we use $y_x$ to denote $Y_x=y$, $X$ stands for treatment, and $Y$ stands for effect. The experimental distributions in this paper are those in the form of the causal effects, $P(y_x)$, and the observational distributions in this paper are those joint probability function $P(x,y)$. 

Three basic probabilities of causation are defined as follows:
\begin{definition}[Probability of necessity (PN)]
Let $X$ and $Y$ be two binary variables in a causal model $M$, let $x$ and $y$ stand for the propositions $X=true$ and $Y=true$, respectively, and $x'$ and $y'$ for their complements. The probability of necessity is defined as the expression \cite{pearl1999probabilities}\\
\begin{eqnarray}
\text{PN} & \delequal & P(Y_{x'}=false|X=true,Y=true)\nonumber \\
& \delequal & P(y'_{x'}|x,y) \nonumber
\label{pn}
\end{eqnarray}
\end{definition}
% \par
% In other words, PN stands for the probability that event $y$ would not have occurred in the absence of event $x$, given that $x$ and $y$ did in fact occur.

% Note that lower case letters (e.g., $x,y$) stand for propositions (or events). PN has applications in epidemiology, legal reasoning, and artificial intelligence. Epidemiologists have long been concerned with estimating the probability that a certain case of disease is attributable to a particular exposure, which is normally interpreted counterfactually as ``the probability that disease would not have occurred in the absence of exposure, given that disease and exposure did in fact occur." This counterfactual notion is also used frequently in lawsuits, where legal responsibility is at the center of contention.
% \vspace{10pt}
\begin{definition}[Probability of sufficiency (PS)] \cite{pearl1999probabilities}
\begin{eqnarray}
\text{PS} \delequal P(y_x|y',x') \nonumber
\label{ps}
\end{eqnarray}
\end{definition}

% PS finds applications in policy analysis, artificial intelligence, and psychology.
% A policy maker may well be interested in the dangers that a certain exposure may present to the healthy population \cite{khoury1989measurement}. Counterfactually, this notion is expressed as the ``probability that a healthy unexposed individual would have gotten the disease had he/she been exposed." In psychology, PS serves as the basis for Cheng's \cite{cheng1997covariation} causal power theory \cite{glymour2013psychological}, which attempts to explain how humans judge causal strength among events. In artificial intelligence, PS plays a major role in the generation of explanations \cite{pearl2009causality}.
% \vspace{10pt}
\begin{definition}[Probability of necessity and sufficiency (PNS)] \cite{pearl1999probabilities}
\begin{eqnarray}
\text{PNS}\delequal P(y_x,y'_{x'}) \nonumber
\label{pns}
\end{eqnarray}
\end{definition}
\par
PNS stands for the probability that $y$ would respond to $x$ both ways, and therefore measures both the sufficiency and necessity of $x$ to produce $y$.

The tight bounds of PNS, PN and PS derived by Tian and Pearl \cite{tian2000probabilities} are then in the following forms:\\

\begin{eqnarray}
\max \left \{
\begin{array}{cc}
0, \\
P(y_x) - P(y_{x'}), \\
P(y) - P(y_{x'}), \\
P(y_x) - P(y)\\
\end{array}
\right \}
\le \text{PNS}
 \le \min \left \{
\begin{array}{cc}
 P(y_x), \\
 P(y'_{x'}), \\
P(x,y) + P(x',y'), \\
P(y_x) - P(y_{x'}) + P(x, y') + P(x', y)
\end{array} 
\right \}
\label{pnsub}
\end{eqnarray}

\begin{eqnarray*}
\max \left \{
\begin{array}{cc}
0, \\
\frac{P(y)-P(y_{x'})}{P(x,y)}
\end{array} 
\right \}
\le \text{PN} \le
\min \left \{
\begin{array}{cc}
1, \\
\frac{P(y'_{x'})-P(x',y')}{P(x,y)} 
\end{array}
\right \}
\label{pnub}
\end{eqnarray*}

\begin{eqnarray*}
\max \left \{
\begin{array}{cc}
0, \\
\frac{P(y')-P(y'_{x})}{P(x',y')}
\end{array} 
\right \}
\le \text{PS} \le
\min \left \{
\begin{array}{cc}
1, \\
\frac{P(y_{x})-P(x,y)}{P(x',y')} 
\end{array}
\right \}
\label{psub}
\end{eqnarray*}

To obtain bounds for a specific subpopulation, defined by a set $C$ of characteristics, the expressions above should be modified by conditioning each term on $C=c$. Therefore, if the experimental and observational distributions are available for every subpopulation, we are able to estimate the probabilities of causation of every subpopulation (here, we reviewed Tian-Pearl bounds for PNS, PS, and PN; Li and Pearl provided bounds for all types of probabilities of causation in \cite{li:pea-r516}). However, in practice, some subpopulations have no adequate data (due to finite population data) to estimate their experimental and observational distributions. In this paper, we propose a machine learning model that takes the bounds of the probabilities of causation (i.e., those bounds that have adequate data to estimate) as the label and provides the estimations of the probabilities of causation of all rest subpopulations.

\section{Causal Model}
\label{causalmodel}
In order to verify the accuracy of the learned bounds of PNS, we must first understand the data generating process to have the true PNS value and its bounds. The model we are using is shown in Figure \ref{causal1} (the coefficients in SCM are randomly generated as in \cite{li:pea22-r518}, see the appendix for the detail), where $X$ is a binary treatment, $Y$ is a binary effect, and $Z$ is a set of $20$ independent binary features (say $Z_1,...,Z_{20}$). The structural equations are as follow (for simplicity reason, we let $x=1,x'=0$, and $y=1,y'=0$):
\begin{figure}
            \centering
            \begin{tikzpicture}[->,>=stealth',node distance=2cm,
              thick,main node/.style={circle,fill,inner sep=1.5pt}]
              \node[main node] (1) [label=above:{$Z$}]{};
              \node[main node] (3) [below left =1cm of 1,label=left:$X$]{};
              \node[main node] (4) [below right =1cm of 1,label=right:$Y$] {};
              \path[every node/.style={font=\sffamily\small}]
                (1) edge node {} (3)
                (1) edge node {} (4)
                (3) edge node {} (4);
            \end{tikzpicture}
            \caption{The Causal Model, where $X$, $Y$ and $Z$ are binary treatment, binary effect, and $20$ independent binary features, respectively.}
            \label{causal1}
\end{figure}
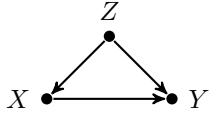

\begin{eqnarray*}
Z_i &=& U_{Z_i} \text{ for } i \in \{1,...,20\},\\
%X &=& 1 \text{ if } M_X +U_X > 3,\\
X&=&f_X(M_X,U_X) =
\left \{
\begin{array}{cc}
1& \text{ if } M_X +U_X > 0.5, \\
0& \text{otherwize},
\end{array}
\right \}\\
%Y &=& 0.867991236792X+M_Y+U_Y,\\
Y&=&f_Y(X,M_Y,U_Y) =
\left \{
\begin{array}{cc}
1& \text{ if } 0< CX+M_Y+U_Y < 1\\
1& \text{ if } 1 < CX+M_Y+U_Y < 2,\\
% 1& \text{ if } 1 < CX+M_Y+U_Y < 2, \\
0& \text{otherwize},
\end{array}
\right \}\\
&&\text{where, } U_{Z_i}, U_X, U_Y \text{ are binary exogenous variables with Bernoulli distributions,}\\ &&\text{C is a constant, and }M_X,M_Y\text{ are linear combinatations of }Z_{1},...,Z_{20}.
\end{eqnarray*}
The value of $C,M_X,M_Y$ and the distributions of $U_X,U_Y,U_{Z_1},...,U_{Z_{20}}$ for the model are provided in the appendix.

\section{Data Generating Process}
Based on the model defined in the last section, there are $20$ binary features. We made $15$ of them observable and $5$ of them unobservable. All exogenous variables are also made unobservable. We then have $2^{15}$ observed subpopulations.
\subsection{Informer Data}
The informer data must know the actual bounds of PNS of each subpopulation for comparison purposes. From the structural equation given in last section, the value of $X$ is determined by $U_X$ and $M_x$ (denoted by $f_X(M_X,U_X)$) and the value of $Y$ is determined by $X$, $M_Y$, and $U_Y$ (denoted by $f_Y(X,M_Y,U_Y)$). If all $20$ binary features are observable, then for a particular features $z=(z_1,...,z_{20})$, $M_X$ and $M_Y$ are fixed (denoted by $M_X(z)$ and $M_Y(z)$), then the PNS, experimental distribution, and observational distribution of this set of features are
\begin{eqnarray*}
PNS(z) &=& P(Y=0_{X=0}, Y=1_{X=1}|z)\\
&=& P(U_Y=0)*T_0 + P(U_Y=1)*T_1, \\
\text{where}, &T_0& =
\left \{
\begin{array}{cc}
1& \text{ if } Y(0,M_Y(z),0)=0 \text{ and } Y(1,M_Y(z),0)=1, \\
0& \text{otherwize},
\end{array}
\right\},\\
&T_1& =
\left \{
\begin{array}{cc}
1& \text{ if } Y(0,M_Y(z),1)=0 \text{ and } Y(1,M_Y(z),1)=1, \\
0& \text{otherwize}
\end{array}
\right \}.
\end{eqnarray*}
\begin{eqnarray*}
&&P(Y=1|do(X),z)\\
&=& P(U_Y=0)*Y(X,M_Y(z),0) + P(U_Y=1)*Y(X,M_Y(z),1).
\end{eqnarray*}
\begin{eqnarray*}
&&P(Y=1|X,z)\\
&=& P(U_X=0)*P(U_Y=0)*Y(X(M_X(z),0),M_Y(z),0)+\\
&&P(U_X=0)*P(U_Y=1)*Y(X(M_X(z),0),M_Y(z),1)) +\\ &&P(U_X=1)*P(U_Y=0)*Y(X(M_X(z),1),M_Y(z),0)) +\\ &&P(U_X=1)*P(U_Y=1)*Y(X(M_X(z),1),M_Y(z),1)).
\end{eqnarray*}
We assumed $15$ of the features are observable (say $Z_1,...,Z_{15}$), which means each subpopulation $c=(z_1,...,z_{15})$ consists $32$ sets of $20$ binary features (say $s_{0}=(z_1,...,z_{15},0,0,0,0,0), s_{1}=(z_1,...,z_{15},0,0,0,0,1), s_{2}=(z_1,...,z_{15},0,0,0,1,0), ...,s_{31}=(z_1,...,z_{15},1,1,1,1,1)$), then we have the PNS, experimental distribution, and observational distribution of all observed subpopulations are as follow:
\begin{eqnarray*}
PNS(c) &=& P(Y=0_{X=0}, Y=1_{X=1}|c)\\
&=& P(s_{0})/P(c)PNS(s_{0})+P(s_{1})/P(c)PNS(s_{1})+\\
&&P(s_{2})/P(c)PNS(s_{2})+...+P(s_{31})/P(c)PNS(s_{31})\\
&=& P(Z_{16}=0)P(Z_{17}=0)P(Z_{18}=0)P(Z_{19}=0)P(Z_{20}=0)PNS(s_{0})+\\
&&P(Z_{16}=0)P(Z_{17}=0)P(Z_{18}=0)P(Z_{19}=0)P(Z_{20}=1)PNS(s_{1})+...+\\
&&P(Z_{16}=1)P(Z_{17}=1)P(Z_{18}=1)P(Z_{19}=1)P(Z_{20}=1)PNS(s_{31}).
\end{eqnarray*}
\begin{eqnarray*}
&&P(Y=1|do(X),c)\\
&=& P(Z_{16}=0)P(Z_{17}=0)P(Z_{18}=0)P(Z_{19}=0)P(Z_{20}=0)P(Y=1|do(X),s_{0})+\\
&& P(Z_{16}=0)P(Z_{17}=0)P(Z_{18}=0)P(Z_{19}=0)P(Z_{20}=1)P(Y=1|do(X),s_{1})+\\
&& P(Z_{16}=0)P(Z_{17}=0)P(Z_{18}=0)P(Z_{19}=1)P(Z_{20}=0)P(Y=1|do(X),s_{2})+...+\\
&& P(Z_{16}=1)P(Z_{17}=1)P(Z_{18}=1)P(Z_{19}=1)P(Z_{20}=1)P(Y=1|do(X),s_{31}).
\end{eqnarray*}
\begin{eqnarray*}
&&P(Y=1|X,c)\\
&=& P(Z_{16}=0)P(Z_{17}=0)P(Z_{18}=0)P(Z_{19}=0)P(Z_{20}=0)P(Y=1|X,s_{0})+\\
&& P(Z_{16}=0)P(Z_{17}=0)P(Z_{18}=0)P(Z_{19}=0)P(Z_{20}=1)P(Y=1|X,s_{1})+\\
&& P(Z_{16}=0)P(Z_{17}=0)P(Z_{18}=0)P(Z_{19}=1)P(Z_{20}=0)P(Y=1|X,s_{2})+...+\\
&& P(Z_{16}=1)P(Z_{17}=1)P(Z_{18}=1)P(Z_{19}=1)P(Z_{20}=1)P(Y=1|X,s_{31}).
\end{eqnarray*}
The informer view of the bounds of $PNS(c)$ (i.e., true bounds) could be obtained using Equation \ref{pnsub} and above observational and experimental distributions.

\subsection{Experimental Sample}
Here is how we collected $5000000$ experimental samples for training purposes. From the causal model in Section \ref{causalmodel}, an individual is determined by $U_X$, $U_Y$, and $U_{Z_1},...,U_{Z_{20}}$. Therefore, we first randomly generated $(U_X,U_Y,U_{Z_1},...,U_{Z_{20}})$ using their distributions; then we randomly generated $X$ using $Bernoulli(0.5)$; the value of $Y$ is then $f_Y(X,M_Y,U_Y)$. We then obtained a experimental sample, $(U_{Z_1},U_{Z_2},...,U_{Z_{15}},X,Y)$ (since $15$ features are observable and identical to $U_{Z_i}$).
\subsection{Observational Sample}
Here is how we collected the $5000000$ observational samples for training purposes. Similarly to experimental data, we first randomly generated $(U_X,U_Y,U_{Z_1},...,U_{Z_{20}})$ using their distributions; the value of $X$ is then $f_X(M_X,U_X)$; the value of $Y$ is then $f_Y(X,M_Y,U_Y)$. We then collect a observational sample, $(U_{Z_1},U_{Z_2},...,U_{Z_{15}},X,Y)$ (since $15$ features are observable and identical to $U_{Z_i}$).

\section{Machine Learning Model}
\subsection{Features and Label}
The training features are $15$ observed features. We obtain the training label as follows: if a given set of features appeared more than $1300$ times (note we use the number $1300$ to have a precise estimation of PNS suggested by Li, Mao, and Pearl \cite{li:pea22-r518}) in those $5000000$ experimental samples and appeared more than $1300$ times in those $5000000$ observational samples, Frequentist will be used to estimate the experimental and observational distributions. We will then apply Equation \ref{pnsub} to obtain lower bound and upper bound for this set of features (lower bound and upper bound are the labels). We totally have $529$ set of features of both lower and upper bounds for training purposes. The $529$ sets of features are then split into $423$ for the training set and $106$ for the testing set.

\subsection{Learning}

We used a simple fully-connected neural network to predict the lower and upper bounds from the feature set. Specifically, we use four multilayer perceptron (MLP) layers. ReLU is used for activation in the leading three layers while Sigmoid function is used for the output layer. We set the embeddings dimension as 128 for all layers, and train the model for 600 iterations with learning rate 0.01. Our experiments are done at an AWS p3.2xlarge instance.

\section{Experimental Results}
\begin{figure}
\centering
\begin{subfigure}[b]{0.49\textwidth}
\includegraphics[width=0.99\textwidth]{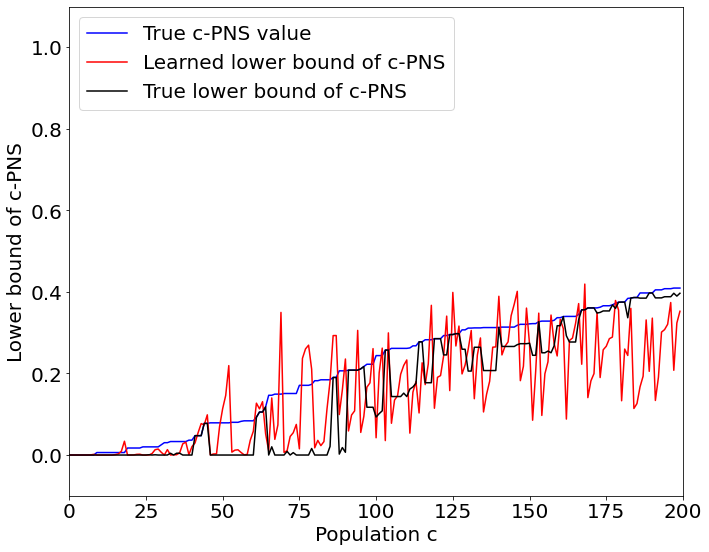}
\caption{Lower bound of the PNS for subpopulations. These $200$ subpopulations are randomly selected from $32768$ subpopulations.}
\label{res1}
\end{subfigure}
\hfill
\begin{subfigure}[b]{0.49\textwidth}
\includegraphics[width=0.99\textwidth]{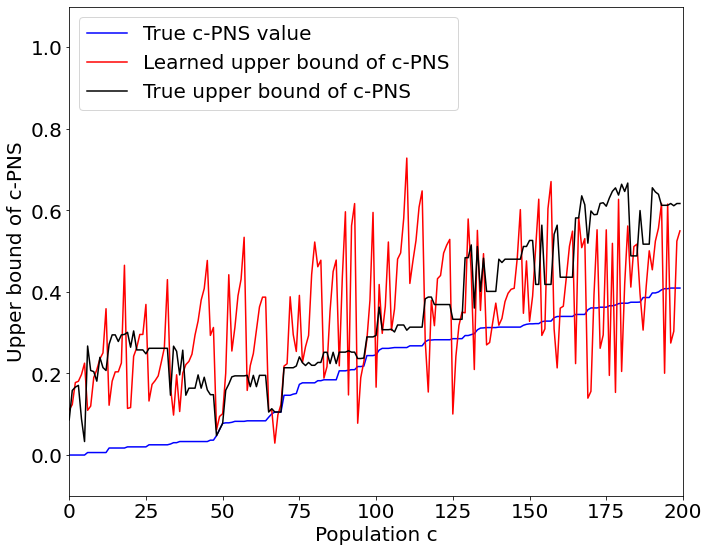}
\caption{Upper bound of the PNS for subpopulations. These $200$ subpopulations are randomly selected from $32768$ subpopulations.}
\label{res2}
\end{subfigure}
\caption{The learned bounds of PNS for subpopulations compared to the true bounds of PNS.}
\label{f1}
\end{figure}

We randomly selected $200$ subpopulations (among $32768$) and compared their learned bounds of PNS with the true bounds of PNS computed from the informer data. The results are shown in Figure~\ref{f1}. The learned bounds of PNS for subpopulations are a good fit for the true PNS bounds. The average error of the learned lower bound among $32768$ subpopulations is $0.0775$, and the average error of the learned upper bound among $32768$ subpopulations is $0.1371$. They are both acceptable errors given we have only $423$ training size to learn $32768$ subpopulations.

% \subsection{Lower Bounds of the PNS}
% \subsection{Upper Bounds of the PNS}

\section{Discussion}
We demonstrated that the probabilities of causation for subpopulations could be learned from finite population data. However, we must discuss some properties of our proposed method further.

First, we applied the machine learning model to learn the bounds of PNS in this paper. This method can be applied to any probability of causation. Machine learning provides the ability to learn the probabilities of causation of subpopulations such that there is insufficient data to estimate the experimental and observational distributions. However, the key is that the label of the machine learning model should be the bounds of the probabilities of causation rather than experimental and observational data. Therefore, the size of the training set is no longer $5000000$ observational and experimental samples; it is the observational and experimental distributions and the bounds obtained from the distributions (in the example we had in the last section, our training data set are the size of $529$).

Second, we applied the simplest machine learning model. However, the key is the framework we proposed. The data generating process is also available; this is the first publicly available data generating process that can test approaches for counterfactual learning. Researchers who are familiar with fancy machine learning models are welcome to apply other machine learning models to this dataset.

Third, this work can be extended to unit selection problems \cite{li:pea22-r517,li:pea19-r488} because the unit selection problems are a linear combination of the probabilities of causation. One can also apply this machine learning framework to predict the experimental distributions \cite{li2022bounds,pearl1995causal} using observational data.

\section{Conclusion}
We demonstrated how to obtain bounds of probabilities of causation for subpopulations using finite population data. We proposed a machine learning framework to deliver reasonable estimations. Experiments showed that the probabilities of causation defined by SCM are learnable with proper labels. Data-generating processes are also available for future machine learning models.

\section*{Acknowledgements}
This research was supported in parts by grants from the National Science
Foundation [\#IIS-2106908 and \#IIS-2231798], Office of Naval Research [\#N00014-21-1-2351], and Toyota Research Institute of North America
[\#PO-000897].
\bibliographystyle{plain}
\bibliography{neurips_2022.bib}

\newpage
\appendix

\section{Appendix}

\subsection{The Causal Model}
The first model in \cite{li:pea22-r518} are used where the coefficients for $M_X,M_Y$ and $C$ were uniformly generated from $[-1,1]$, and the Bernoulli distribution parameters were uniformly generated from $[0,1]$. The detailed model is as follows:
\begin{eqnarray*}
Z_i &=& U_{Z_i} \text{ for } i \in \{1,...,20\},\\
%X &=& 1 \text{ if } M_X +U_X > 3,\\
X&=&f_X(M_X,U_X) =
\left \{
\begin{array}{cc}
1& \text{ if } M_X +U_X > 0.5, \\
0& \text{otherwize},
\end{array}
\right \}\\
%Y &=& 0.867991236792X+M_Y+U_Y,\\
Y&=&f_Y(X,M_Y,U_Y) =
\left \{
\begin{array}{cc}
1& \text{ if } 0< CX+M_Y+U_Y < 1 \text{ or } 1 < CX+M_Y+U_Y < 2,\\
% 1& \text{ if } 1 < CX+M_Y+U_Y < 2, \\
0& \text{otherwize},
\end{array}
\right \}\\
&&\text{where, } U_{Z_i}, U_X, U_Y \text{ are binary exogenous variables with Bernoulli distributions.}
s.t., \\
&&U_{Z_1} \sim \text{Bernoulli}(0.352913861526), U_{Z_2} \sim \text{Bernoulli}(0.460995855543),\\
&&U_{Z_3} \sim \text{Bernoulli}(0.331702473392), U_{Z_4} \sim \text{Bernoulli}(0.885505026779),\\
&&U_{Z_5} \sim \text{Bernoulli}(0.017026872706), U_{Z_6} \sim \text{Bernoulli}(0.380772701708),\\
&&U_{Z_7} \sim \text{Bernoulli}(0.028092602705), U_{Z_8} \sim \text{Bernoulli}(0.220819399962),\\
&&U_{Z_9} \sim \text{Bernoulli}(0.617742227477), U_{Z_{10}} \sim \text{Bernoulli}(0.981975046713),\\
&&U_{Z_{11}} \sim \text{Bernoulli}(0.142042291381), U_{Z_{12}} \sim \text{Bernoulli}(0.833602592350),\\
&&U_{Z_{13}} \sim \text{Bernoulli}(0.882938907115), U_{Z_{14}} \sim \text{Bernoulli}(0.542143191999),\\
&&U_{Z_{15}} \sim \text{Bernoulli}(0.085023436884), U_{Z_{16}} \sim \text{Bernoulli}(0.645357252864),\\
&&U_{Z_{17}} \sim \text{Bernoulli}(0.863787135134), U_{Z_{18}} \sim \text{Bernoulli}(0.460539711624),\\
&&U_{Z_{19}} \sim \text{Bernoulli}(0.314014079207), U_{Z_{20}} \sim \text{Bernoulli}(0.685879388218),\\
&&U_{X} \sim \text{Bernoulli}(0.601680857267), U_{Y} \sim \text{Bernoulli}(0.497668975278),\\
&&C=-0.77953605542,
\end{eqnarray*}
\begin{eqnarray*}
&M_X& =
\begin{bmatrix}
Z_1~Z_2~...~Z_{20}
\end{bmatrix}\times
\begin{bmatrix}
0.259223510143\\ -0.658140989167\\ -0.75025831768\\ 0.162906462426\\ 0.652023463285\\ -0.0892939586541\\ 0.421469107769\\ -0.443129684766\\ 0.802624388789\\ -0.225740978499\\ 0.716621631717\\ 0.0650682260309\\ -0.220690334026\\ 0.156355773665\\ -0.50693672491\\ -0.707060278115\\ 0.418812816935\\ -0.0822118703986\\ 0.769299853833\\ -0.511585391002
\end{bmatrix},
M_Y =
\begin{bmatrix}
Z_1~Z_2~...~Z_{20}
\end{bmatrix}\times
\begin{bmatrix}
-0.792867111918\\ 0.759967136147\\ 0.55437722369\\ 0.503970540409\\ -0.527187144651\\ 0.378619988091\\ 0.269255196301\\ 0.671597043594\\ 0.396010142274\\ 0.325228576643\\ 0.657808327574\\ 0.801655023993\\ 0.0907679484097\\ -0.0713852594543\\ -0.0691046005285\\ -0.222582013343\\ -0.848408031595\\ -0.584285069026\\ -0.324874831799\\ 0.625621583197
\end{bmatrix}
\end{eqnarray*}

\end{document}